\documentclass[final,5p,times,twocolumn,article]{elsarticle}

\usepackage{amssymb}
\usepackage{xcolor}
\usepackage{tabularx}
\usepackage{lineno}
\usepackage{multirow}
\usepackage{hyperref}       
\hypersetup{
    colorlinks=true,
    linkcolor=blue,
    filecolor=magenta,      
    urlcolor=magenta,
    citecolor=blue,
}
\usepackage{booktabs}
\usepackage{amsmath}

\journal{Cities}

\begin{document}

\begin{frontmatter}



\title{Multi‑Modal Feature Fusion for Spatial Morphology Analysis of Traditional Villages via Hierarchical Graph Neural Networks} 

\author{Jiaxin Zhang $^{1}$, Zehong Zhu $^{2}$, Junye Deng $^{2}$, Yunqin Li $^{1,}$* and Bowen Wang $^{3}$\\
Architecture and Design College, Nanchang University\\
SANKEN, The University of Osaka\\
{\tt\small $^1$ \{jiaxin.arch, liyunqin\}@ncu.edu.cn} \\
{\tt\small $^2$\{6011121021, 6011121055\}@email.ncu.edu.cn}\\
{\tt\small $^3$wang@im.sanken.osaka-u.ac.jp}\\
}




\begin{abstract}
Villages areas hold significant importance in the study of human-land relationships. However, with the advancement of urbanization, the gradual disappearance of spatial characteristics and the homogenization of landscapes have emerged as prominent issues. Existing studies primarily adopt a single-disciplinary perspective to analyze villages spatial morphology and its influencing factors, relying heavily on qualitative analysis methods. These efforts are often constrained by the lack of digital infrastructure and insufficient data. To Address the current research limitations, this paper proposes a Hierarchical Graph Neural Network (HGNN) model that integrates multi-source data to conduct an in-depth analysis of villages spatial morphology. The framework includes two types of nodes—input nodes and communication nodes—and two types of edges—static input edges and dynamic communication edges. By combining Graph Convolutional Networks (GCN) and Graph Attention Networks (GAT), the proposed model efficiently integrates multimodal features under a two-stage feature update mechanism. Additionally, based on existing principles for classifying villages spatial morphology, the paper introduces a relational pooling mechanism and implements a joint training strategy across 17 subtypes. Experimental results demonstrate that this method achieves significant performance improvements over existing approaches in multimodal fusion and classification tasks. Additionally, the proposed joint optimization of all sub-types lifts mean accuracy/F1 from \textbf{0.71~/~0.83} (independent models) to \textbf{0.82~/~0.90}, driven by a \textbf{6\,\%} gain for parcel tasks. Our method providing scientific evidence for exploring villages spatial patterns and generative logic.
\end{abstract}



\begin{keyword}
Multi-source Data \sep Traditional Villages \sep Spatial Morphology Analysis \sep Information Fusion \sep Graph Neural Networks 



\end{keyword}

\end{frontmatter}



\section{Introduction}
Traditional village areas represent a complex, self-organizing system that continuously evolves through interactions between human societies and their surrounding environments. As the central spatial unit of villages geography research, they play a crucial role in studies of human-land interrelations \cite{Fujii2003,Huang2024,Hualou2013}. Over the long course of historical evolution, villages areas in China have developed regionally distinctive spatial morphologies with pronounced local characteristics. Such evolution bears significant implications for the transmission of China’s tangible and intangible cultural heritage, the continuity of national spirit, and sustainable development \cite{Guo2016,Deng2024,Li2020a}. On one hand, these spatial morphologies differ because villages areas are formed by self-organization and environmental adaptation influenced by geographical features. On the other hand, villages development is shaped by various sociocultural factors, such as demographic economics, cultural identity, and the continuation of traditions \cite{Tan2021,Yao2023}. The spatial form, geographic attributes, and sociocultural elements, along with their interconnections, jointly account for regional differences among villages areas \cite{Ge2024,Chen2022}. With the ongoing process of urbanization, villages regions face a stark conflict between economic growth and the preservation of cultural heritage. The gradual disappearance of villages characteristics and the homogenization of landscapes have become prominent negative factors in both urban and villages human environments \cite{Liu2022,zhang2024archgpt}. Behind this phenomenon lies an insufficient understanding of traditional villages culture and a lack of in-depth knowledge concerning villages spatial characteristics and cultural connotations \cite{Yan2017}. Consequently, establishing an understanding of villages space and its governing spatial laws is urgently needed for exploring new models of villages revitalization and for tailoring villages landscape designs to local conditions. A thorough investigation into the spatial morphologies of different villages types and their underlying mechanisms constitutes a critical means of objectively understanding villages development and spatial patterns.

Currently, various disciplines each emphasize different perspectives and methods regarding the factors that affect villages space. From the perspective of disciplinary research, geography primarily examines the spatial distribution of entire villages from a macro viewpoint, quantifying and characterizing geographic spatial relationships to elucidate distribution patterns in various regional types \cite{Bian2022, Feng2023, Zhang2024}. Architecture and planning disciplines often employ typological methods to summarize spatial morphological features at the meso- and micro-scales, such as villages landscape layouts \cite{Sa2024,zhang2024archgpt}, village textures \cite{Hou2021}, street and alley transportation \cite{Lin2024}, and architectural forms \cite{Afnarius2020,wang2024improving}. These studies also identify the factors that contribute to the formation of different morphological types. Some researchers have quantitatively described morphological changes in settlements by measuring the geometric properties of villages—such as boundary shape, building density, courtyard space ratio, fractal dimension, and shape analysis indices—to achieve precise and scientific descriptions of settlement morphologies \cite{chen2023x}. Methodologically, due to the limited digital infrastructure in villages areas and the scarcity of open-source data and data-related technologies, research on villages spatial morphology has largely relied on qualitative analysis methods, including surveys, interviews, and data compilation \cite{Chai2022,Ren2022,wang2023match}. With advances in computational technology, quantitative analytical methods are increasingly being adopted, facilitating the discovery of new, scientifically validated rules through the application of methods such as morphological clustering, kernel density estimation, and space syntax \cite{Zhu2023}. Beyond data-driven quantitative approaches, knowledge-driven AI in specialized fields is on the rise and has begun to be incorporated as a fundamental decision-making tool \cite{Wang2024}.

\begin{figure}[t]
\centering
\includegraphics[width=0.65\linewidth]{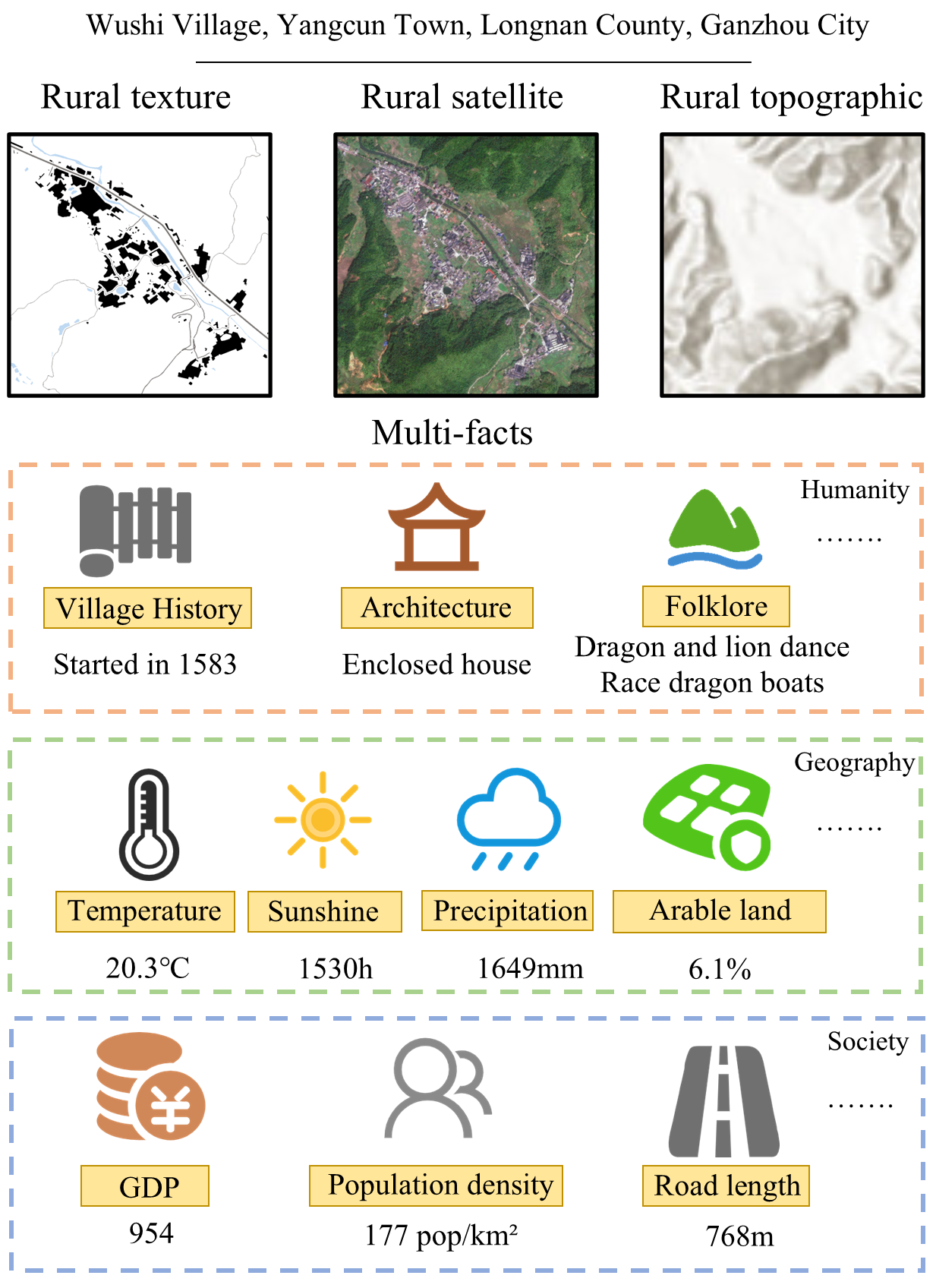}
\caption{Multi-source data used in this research.}
\label{fig:1}
\end{figure}

Machine learning techniques \cite{lecun2015deep,dong2021survey,zhang2024instruct,wang2023learning}, particularly Graph Neural Networks (GNNs) \cite{Sperduti1997,Scarselli2008,wang2024direct,ahmedt2021graph}, have shown tremendous potential in spatial data analysis. Compared to traditional models, GNNs are particularly advantageous when dealing with graph-structured data \cite{Zhou2020,wang2025taming}. Especially Graph Convolutional Networks (GCNs) \cite{kipf2016semi}, leveraging convolutional operations, effectively capturing local and global features, automatically learning robust node representations, and thus reducing the need for extensive feature engineering \cite{Wu2020}. In spatial data analysis, GNNs can exploit the dependencies among nodes and edges, making them well-suited for complex topological structures such as transportation networks, social networks, or molecular graphs \cite{Zhu2022}. Relative to traditional approaches, GNNs not only excel in tasks like node classification and link prediction, but also demonstrate strong modeling capacity by incrementally capturing global structural information through multiple convolutional layers.

Overall, existing studies on villages spatial morphology and its influencing factors often adopt a single perspective, constrained by data availability and lacking consideration of multi-source information. Investigations into spatial laws and mechanisms are typically confined to isolated analyses of a closed typology and have yet to adopt an integrated viewpoint that scientifically recognizes both the spatial laws of villages areas and their generative logic. Methodologically, a heavy reliance on established rules and exclusively geometric attributes of villages space neglects the holistic logical relationships inherent in villages spatial perception. Therefore, to address the aforementioned gaps, this paper proposes a hierarchical GNN model that integrates multi-source data (shown in Figure \ref{fig:1}) to conduct an in-depth analysis of villages spatial morphologies, consolidating multi-source information for analyzing village spatial morphology. This framework features two types of nodes—input nodes and communication nodes—and two types of edges—static input edges and dynamic communication edges. These components cooperate in an iterative process combining GCNs and Graph Attention Networks (GATs) \cite{velickovic2017graph} within a two-stage feature updating scheme. The hierarchical structure exploits the intrinsic characteristics and prior structures of multi-source data, thereby enabling efficient fusion and propagation of multimodal features. Moreover, building on existing principles of villages spatial morphology classification, we develop a relation pooling mechanism and a joint training strategy across all 17 sub-types. Compared to existing methods for multimodal fusion and classification, our approach achieves a substantial performance improvement.

The main contributions of this article include:  
1) Construction of a multi-source dataset for villages spatial morphology analysis;  
2) Development of a hierarchical GNN model for multi-source information fusion;  
3) Introduction of a relation pooling method for sub-type classification based on existing principles.

\section{Related Works}
\subsection{Geographical and Sociocultural Influences on villages Spatial Morphology}
In recent decades, the study of villages spatial morphology has transitioned from traditional morphology to interdisciplinary research, with research content gradually exhibiting diversification trends. Early studies primarily focused on the classification of spatial morphological elements \cite{Peng1992}, the morphological classification of villages areas \cite{Fujii2003}, and spatial transformation and reconstruction \cite{Hualou2013}. With the deepening of research, studies on villages spatial morphology are no longer limited to spatial characteristics alone. Scholars have begun to focus on the social and natural attributes of villages spaces, such as villages revitalization and conservation updates \cite{Deng2024}, villages population mobility \cite{Li2020a}, the socio-cultural structure of villages clans \cite{Wei2015}, and habitat ecological space quality \cite{Li2023}. These studies have broadened the field of villages spatial research, providing multidimensional perspectives for its development.

With the application of quantitative analysis tools in the field of villages space, the evolution of villages spatial morphology and its driving mechanisms have garnered significant attention. Geography and socio-cultural factors are the primary drivers of villages spatial morphology \cite{Chen2022}. For instance, \cite{Li2020} utilized the CLUE-S model to simulate and predict the evolution trends of villages settlements. The study revealed that geographical factors such as energy gradients and surface fragmentation significantly impact villages settlements in mountainous areas. Similarly, \cite{Tan2021} employed GWTR and the geographical detector to unveil the varying impacts of socio-economic factors, such as villages population and mechanical power, under different geographical conditions. \cite{Liu2022} constructed a spatial pedigree of traditional villages, using watersheds as the research scope, highlighting the influence of factors such as waterways, elevation, and roads on villages agglomeration. \cite{Yao2023} demonstrated that socio-economic factors are the main drivers of spatial evolution in villages settlements, while geographical factors have a less significant impact. However, their interaction with socio-economic factors significantly enhances explanatory power. \cite{Liu2023} found that the interaction between historical cultural intensity and population density provides the most significant explanatory power for the spatial distribution of traditional villages. These studies indicate that geographical and socio-cultural factors are intricately intertwined in the evolution of villages spatial morphology, with complex and profound mechanisms. Addressing this complexity necessitates the introduction of advanced analytical frameworks.

With the acceleration of urbanization and regional development, spatial research has increasingly become an essential tool for understanding geographical phenomena and optimizing resource allocation. Traditional statistical methods struggle to capture complex spatial dependencies, while Graph Neural Networks (GNNs) have gained widespread attention for their unique ability to model graph-structured data. As a representative model of GNNs, Graph Convolutional Networks (GCNs) excel in processing spatial adjacency relationships and feature fusion, achieving remarkable results in spatial recognition \cite{Lan2022}, traffic optimization \cite{Ali2022}, and risk prediction \cite{Liang2024}. Urban traffic prediction is one of the key application areas of GCNs, often used for predicting population and traffic flows \cite{Tang2024}. Given that GCN models are better at extracting spatial features from non-Euclidean distance data, many studies have integrated GCN models with temporal models for flow prediction \cite{Zhao2019}. For instance, GCN combined with LSTM or GRU has been applied in taxi demand prediction \cite{Yang2022, Yang2024}. Meanwhile, GCN has demonstrated robust spatial measurement capabilities. \cite{Xue2022} used a GCN-based road centrality metric derived from grid-level road networks in 30 cities to predict spatial homogeneity and employed this metric to assess urban structure and socio-economic performance. \cite{Yan2021} proposed a GCN method for pattern cognition and classification of individual buildings. In contrast to prior studies, we introduce GCN into villages spatial morphology to address the complex mechanisms between villages spatial morphology and influencing factors.

\subsection{Graph Neural Networks}
Sperduti et al. \cite{Sperduti1997} were pioneers in applying neural networks to directed acyclic graphs, laying the groundwork for subsequent studies on graph neural networks (GNNs). The formal notion of GNNs was introduced by Gori et al. \cite{Gori2005} and expanded upon in Scarselli et al. \cite{Scarselli2008}, who provided a framework for learning node representations in graphs. This foundation was further advanced by Gallicchio et al. \cite{Gallicchio2010}, contributing to the development of GNN theory and applications. Motivated by the success of convolutional neural networks (CNNs) in computer vision, researchers sought to redefine convolution operations for graph-structured data, giving rise to convolutional graph neural networks (ConvGNNs). These methods adapt the principles of CNNs to aggregate information from graph structures effectively. By leveraging graph embeddings, ConvGNNs enable the modeling of inputs and outputs that consist of elements and their dependencies. One prominent example is the Graph Convolutional Network (GCN) \cite{kipf2016semi} proposed by Kipf and Welling, which simplifies the convolution operation using spectral methods. This approach reduces computational complexity and addresses overfitting challenges often encountered in graph-based learning. Following this, Velickovic et al. introduced the Graph Attention Network (GAT) \cite{velickovic2017graph}, incorporating the attention mechanism into the message-passing process. GAT employs a self-attention strategy to compute the hidden states of each node, focusing on its most relevant neighbors. The evolution of GNNs continues to be driven by innovations in both theoretical frameworks and practical applications. Recent advancements have expanded their use in diverse domains, such as natural language processing, recommendation systems, and molecular biology, highlighting the versatility and potential of graph-based deep learning.

With their ability to analyze non-Euclidean spatial data, GNNs exhibit significant potential for processing structured and multi-source information. In critical domains such as medicine \cite{wang2024direct,ahmedt2021graph}, IoT \cite{liang2022survey}, and urban analysis \cite{li2022graph,sharma2023graph}, GNNs are increasingly demonstrating their versatility and applicability. For instance, Li et al. \cite{li2022graph} explored the use of GNNs in various urban intelligence applications, including modeling urban road network structures \cite{wegner2015road}, analyzing gas supply pipelines \cite{holmberg2022jet}, and partitioning cities into functional regions \cite{rong2023goddag,zhang2020local}. Similarly, Silva et al. \cite{silva2024using} applied GNNs to predict local cultural attributes using a large-scale public dataset, demonstrating the value of structural connectedness in forecasting neighborhood characteristics. While their objective aligns with ours—leveraging multi-source data to forecast local labels—their analysis was limited to non-image data, which constrained the breadth and generalizability of their findings. GraphSAGE \cite{lei2024predicting} inferred building attributes by proposing a more spatially explicit representation of urban features as predictors. Using a graph-structured approach, they incorporated the spatial relationships between buildings and street-level urban objects (such as nearby amenities and urban furniture), along with streetscape images as inputs, to predict building attributes. Unlike previous work, our task provides a more complex multi-source fusion task and applies a hierarchy GNN to explore the potential relations among different sources.

\begin{figure}[t]
\centering
\includegraphics[width=0.95\columnwidth]{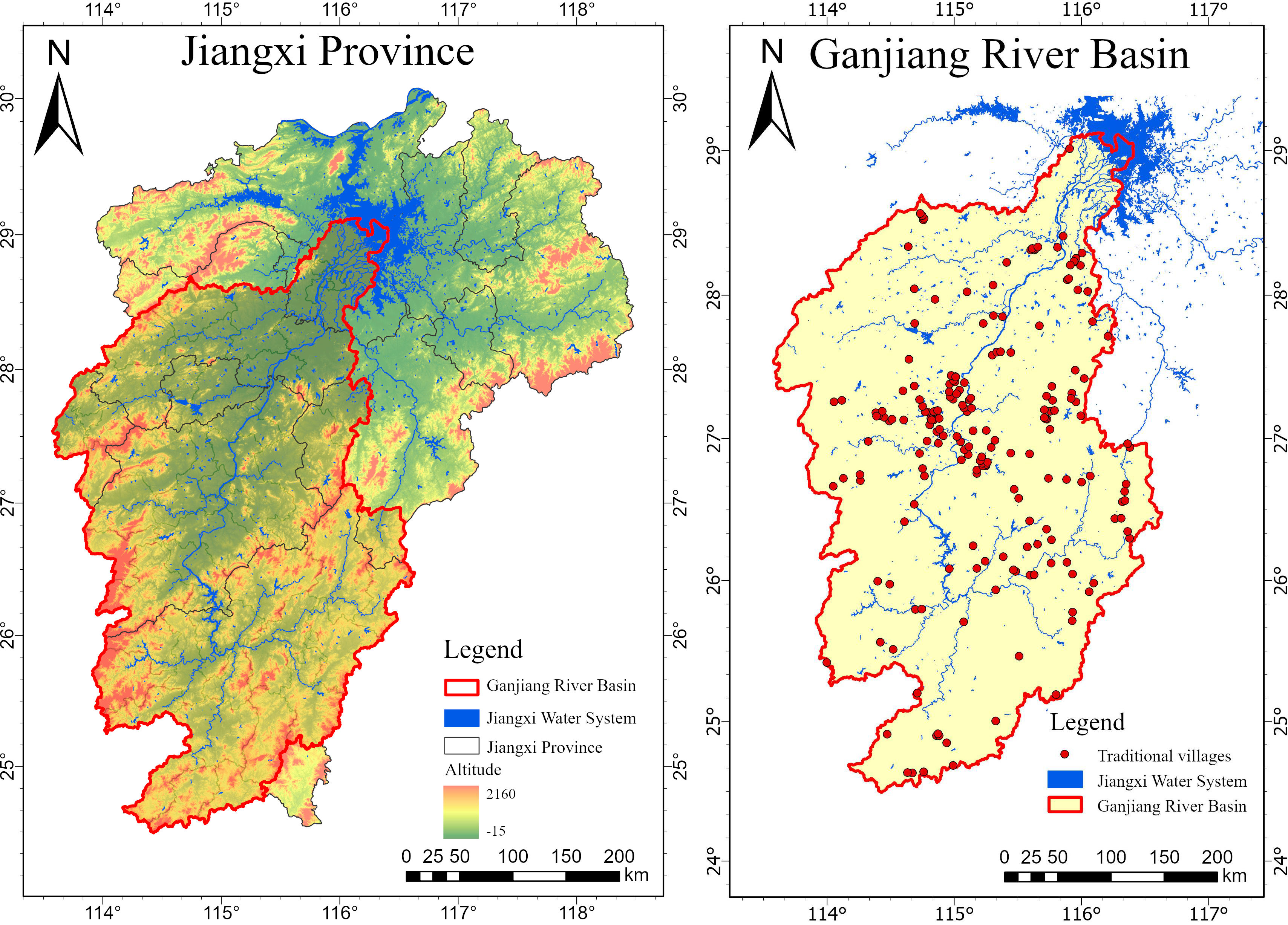}
\caption{The study area in this research.}
\label{fig:study}
\end{figure}

\section{Methods}
\subsection{Dataset Construction}
We collected multi-source data from 583 villages in Jiangxi Province, Ganjiang River, each accompanied by multi-source information and corresponding spatial morphology types (the study area is shown in Figure \ref{fig:study}). The dataset includes three types of input data: images, text, and multi-source factors, as well as three types of predicted data: settlement landscape spatial structure (S), settlement and parcel morphological patterns (V), and settlement road network patterns (R). This dataset provides a rich foundation for studying village spatial morphology and its influencing factors.

We collected multi-source input data with the sources as follows:

\textbf{Images}: Villages texture images, villages satellite maps, and villages topographic maps were obtained from the National Geospatial Information Public Service Platform.

\textbf{Text}: Villages textual data were sourced from government official websites, the Chinese Traditional Villages Digital Museum, the Yunshang Ancient Villages Digital Museum, and Baidu Baike. It introduce the brief information of the villages.

\textbf{Independent Multi-facts}: This is divided into three categories, including: (1) \textbf{Humanities}, further divided into 2 subcategories and 5 sub-subcategories, namely Folklore, Architecture, Clan, Village History, and Famous People, all sourced from government websites and local county chronicles. (2) \textbf{Geography}, further divided into 5 subcategories and 11 sub-subcategories, including Administrative divisions and Coordinates from the China Traditional Village Catalog, Altitude and Elevation from the Geospatial Data Cloud, Density of the water network and Distance to the nearest waterbody from OpenStreetMap, Precipitation, Sunshine, and Temperature from the National Tibetan Plateau Scientific Data Center, and Percentage of arable land and Land Normalized Vegetation Index from the National Ecosystem Data Center. (3) \textbf{Society}, further divided into 4 subcategories and 9 sub-subcategories, including Population density, Night Light Index, Urbanization rate, Point of Interest, and Gross Domestic Product from the Resource and Environment Science Data Platform, as well as Road length, Distance to ancient stage road (Existing studies indicate that the historical transportation framework still exerts a profound influence on contemporary village patterns \cite{Ma2023, Lin2024}.), Texture Dispersion, and Texture Density, all derived from processed village texture vector maps provided by the National Geographic Information Public Service Platform, Tianditu. Table \ref{independent} provides a detailed classification and description of each sub-subcategory.

The Images data collected in this study directly reflect the spatial layout, land use, and natural environmental characteristics of the villages, which facilitates quantitative analysis of the interactions between villages and their surrounding environments. The text data provide detailed records of the historical evolution, cultural traditions, and social structures of the villages. These data help to understand the sociocultural driving forces behind the spatial patterns of villages and reveal the profound impact of human activities and cultural inheritance on the evolution of spatial morphology. The independent multi-fact data encompass multi-dimensional statistical information such as population, economy, climate, and land use, providing a basis for quantitatively characterizing the coupling relationships between human activity intensity, socioeconomic development, and the natural environment.

All data underwent systematic preprocessing before being used as input. For image data, we determined the coordinates and spatial extent of each village and standardized the spatial resolution to ensure consistency in spatial reference. Text data were directly crawled from relevant websites. For independent multi-fact data, each village was treated as a unit point. Humanities data, similar to text data, were organized in tabular form, while Geography and Society data were derived from original raster datasets and assigned to each village unit point. All data correspond to the same time point, ensuring spatiotemporal consistency across the dataset.

\begin{table*}[ht!]
  \caption{Description of independent and dependent variables.}
  \centering
  \resizebox{1\textwidth}{!}{
  \begin{tabular}{p{2cm}p{2cm}p{5cm}|p{16cm}}
    \toprule
    \multicolumn{3}{c}{Variable} & \multicolumn{1}{c}{Description}  \\
    \midrule
    \multicolumn{4}{l}{\textbf{Independent Variables}} \\
    \textbf{Humanity} & Culture & Folklore & Stories and legends passed down in the village, reflecting local culture and beliefs. \\  
    & & Architecture & The style and design of buildings in the village, showing local history and way of life. \\
    & & Clan & A family group with a common ancestry, sharing culture and responsibilities. \\
    \cmidrule(lr){2-4}
    & History & Village History & The origin, development, and significant events of the village. \\
    & & Famous People & Important individuals from the village who made contributions in culture, art, or other fields. \\
    \cmidrule(lr){2-4}
    \textbf{Geography} & Location & Administrative divisions & Municipalities, districts and townships where villages are located. \\
    & & Coordinates & Coordinates of the village location in latitude and longitude. \\
    \cmidrule(lr){2-4}
    & Terrain & Altitude & Elevation cell mean. \\
    & & Elevation & Slope cell mean. \\
    \cmidrule(lr){2-4}
    & Hydrology & Density of the water network & Kernel density of water network. \\
    & & Distance to the nearest waterbody & Distance to nearest water network. \\
    \cmidrule(lr){2-4}
    & Climate & Precipitation & Annual precipitation cell mean. \\
    & & Sunshine & Annual sunshine cell mean. \\
    & & Temperature & Annual mean temperature cell mean. \\
    \cmidrule(lr){2-4}
    & Land & Percentage of arable land & Unit average of proportion of arable land. \\
    & & Normalized vegetation index & Unit average of NDVI. \\
    \cmidrule(lr){2-4}
    \textbf{Society} & Population & Population density & Population density unit average. \\
    \cmidrule(lr){2-4}
    & Economy & Night light index & Luminous remote sensing index. \\
    & & Urbanization rate & Urbanization rate of the administrative district in which it is located. \\
    & & Point of interest & Number of POI units. \\
    & & Gross domestic product & GDP cell average. \\
    \cmidrule(lr){2-4}
    & Transportation & Road length & Length of road in cell. \\
    & & Distance to ancient stage road & Distance to nearest ancient stage road. \\
    \cmidrule(lr){2-4}
    & Public Space & Texture dispersion & Village texture density. \\
    & & Texture Density & Village texture discretization. \\
    \midrule
    \multicolumn{4}{l}{\textbf{Dependent variables}} \\
    \multicolumn{2}{l}{\textbf{Settlement landscape spatial structure (S)}} & Typical Pattern Sequence (S1) cls-3 & Refers to the characteristics of the interface ordering of village, mountain, water and field elements centered on the boundary of the settlement entity; calculated by qualitative indicators \\
    && Relationship between village and mountain (S2) cls-4& This refers to the spatial location of the physical boundaries of villages in relation to various types of undulating hills, and is calculated using quantitative indicators, with the symbols $D$ and $C$ for the distance between villages and hills and the degree of hill enclosure, respectively.\\
    && Village water relationship (S3) cls-6& Refers to the geometric relationship between the physical boundaries of the settlement and the river system, calculated as a qualitative indicator.\\
    && Village-field relationship (S4) cls-2& It refers to the positional relationship between the physical boundary of the village and the space of cultivated fields, and is calculated by quantitative indicators, with the symbols $D$ and $C$ for the distance between villages and fields and the degree of enclosure of cultivated fields, respectively.\\
    && Village natural environment type (S5) cls-4& The dominant type of natural environment in the village is defined by comparing the area coverage of mountains, water and fields within the study area, and the symbols for the coverage of mountains, water and fields within the study area are $C$ mountains, $C$ water and $C$ fields, respectively, and the symbols for the coverage of mountains, water and fields within the study area are $C$ mountains, $C$ water and $C$ fields, respectively. \\
    && Water and field pattern (S6) cls-3& Calculations were carried out using a combination of qualitative and quantitative indicators, including a qualitative summary of the abstract pattern of the geometrical form of the settlement field and a quantitative summary of the elements of morphological characteristics, and the symbols for the average length of the side of the field and the average area of the field are D and A, respectively. \\
    \cmidrule(lr){3-4}
    
    \multicolumn{2}{l}{\textbf{Settlement and parcel morphology patterns (V)}} & Shape index of settlement boundary (V1) cls-4& $S = \frac{P}{\left(1.5\sqrt{\lambda} - \sqrt{\lambda} + 1.5\right)} \sqrt{\frac{A\pi}{\pi}}$ where $P$ is the perimeter of the physical boundary of the settlement, $A$ is the area of the boundary shape, and $\lambda$ is the ratio of the long side to the wide side of the outer rectangle of the boundary. \\
    && Degree of fragmentation of settlement boundary (V2) cls-2& $D = \frac{2 \log_{10}\left(\frac{P}{4}\right)}{\log_{10}(A)}$ where $A$ is the perimeter of the boundary of the settlement, $P$ is the area of the boundary shape and $P$ is the area of the boundary shape. \\
    && Degree of dispersion of village plots (V3) cls-2& $S = \frac{1}{n} \sum_{i=1}^{n} \sqrt{(x_i - x)^2 + (y_i - y)^2}$ and $R = \frac{S}{d}$ where $x$ and $y$ are the horizontal and vertical coordinates of the center of the plot system, $x_i$ and $y_i$ are the horizontal and vertical coordinates of the center of the plot system, $S$ is the average distance from the center of the system to each plot point in the system, and $d$ is the length of the diagonal of the boundary of the colony entity. \\
    && Colony density (V4) cls-2& This refers to the proportion of the total area of plots within the settlement boundaries to the area of the settlement study boundary, and is calculated as a quantitative indicator, symbolized by $D$. \\
    && Mean area of cluster plots (V5) cls-2& Refers to the average area of a plot within a settlement, calculated as a quantitative indicator, symbolized as $A$. \\
    \cmidrule(lr){3-4}
    
    \multicolumn{2}{l}{\textbf{Settlement road network pattern (R)}} & Relationship between road network alignment and landscape (R1) cls-2& This refers to the geometric relationship between the road network characterization and the landscape morphology in the settlement, and is calculated by qualitative indicators. \\
    && Relationship between road network alignment and wind direction (R2) cls-3& This refers to the geometric relationship between the alignment of the main road network and the prevailing wind direction within the settlement, and is calculated as a qualitative indicator. \\
    && Road network curvature (R3) cls-3& $C = \frac{D}{D_0}$ where, $D$ is the actual road length in the study area, $D_0$ is the road network length in the topological model.\\
    && Density of road network (R4) cls-2& $D\_d = \frac{D}{A}$, where $D$ is the actual road length within the study area, $A$is the area of the study area.\\
    && Average width of road network at various levels (R5) cls-2& Refers to the average width of the primary and secondary road networks within the settlement, calculated as a quantitative indicator, with the symbols $D$ and $d$ for primary and secondary road networks, respectively.\\
    && Road network structure (R6) cls-3& A qualitative summary of the abstract schematization of the road network within the settlement. \\
    
    \bottomrule
  \end{tabular}
  }
  \label{independent}
\end{table*}

\begin{figure*}[t]
\centering
\includegraphics[width=1\linewidth]{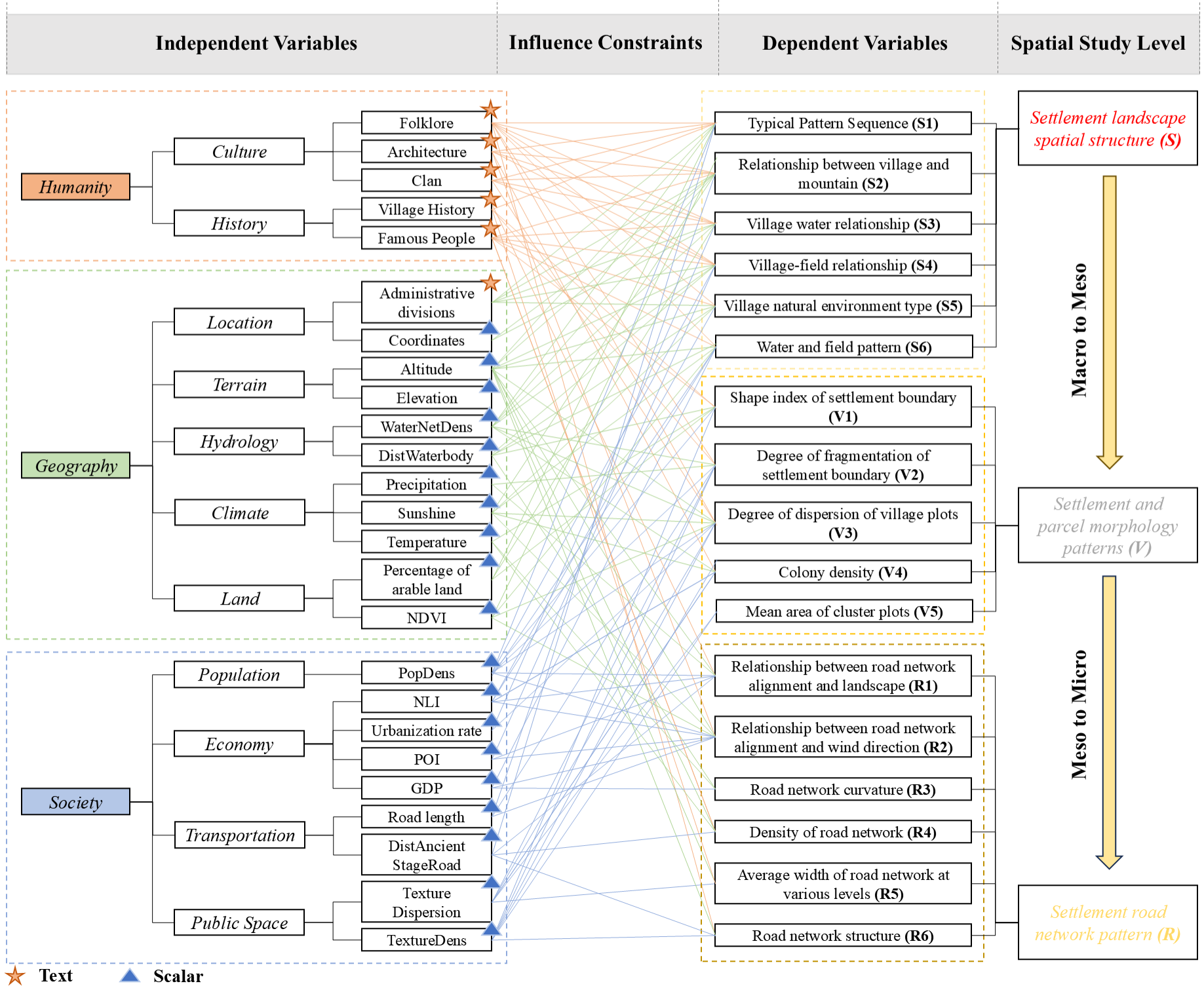}
\caption{Principles of related multi-facts data for a subtype prediction. Note that, image and text data are not associated with any subtype. }
\label{fig:relation}
\end{figure*}

The final prediction of spatial morphology types is a multi-task classification \cite{zhang2021survey,wang2021image}. Morphological studies grounded in geographical theory were significantly advanced by M. R. G. Conzen, who developed the town plan analysis method. This approach, together with subsequent advances in morphological typology, has been extensively applied in urban design and related disciplines. In this context, spatial morphology refers to the form and organizational structure of spatial elements, specifically emphasizing the integration of street systems, plot arrangements, and building coverage within a planning unit \cite{Chen2017TypoMorphological}. Extending this theoretical framework to the study of village spatial morphology, we have identified three groups of predictions encompassing a total of 17 subtypes, which are divided based on research scale into three principal categories: Settlement Landscape Spatial Structure (S), Settlement and Parcel Morphology Patterns (V), and Settlement Road Network Patterns (R). In rural societies, the relatively limited capacity to modify the natural environment means that village site selection, spatial layout, and construction typically prioritize locations favorable for agricultural production and daily living \cite{Cheng2022SpatialGeneMap}. This results in distinctive landscape patterns, making the analysis of S particularly important. In contrast, V and R are analogous to urban plots and street systems. Due to limitations in data availability and research scale, building coverage at the individual building level was not analyzed in this study. The S group is further divided into six subcategories, each with specific classification criteria: Typical Pattern Sequence (S1), including Waterfield wedge village, Waterfield ring village with mountain cluster embracing outside, and Mountain houses integrated with waterfield extending outward; Relationship between Village and Mountain (S2), categorized as Mountain-near village, Mountain-enclosed village, Mountain-pack village, and Village separated from mountains; Village-Water Relationship (S3), including Single-river-adjacent village, Single-river-crossing village, Multi-river-adjacent village, Multi-river-crossing village, Lakeside village, and Lake-encircled village; Village-Field Relationship (S4), classified as Field-near village and Field-enclosed village; Village Natural Environment Type (S5), including Polder field type, Waterfield type, Mountain-water type, and Mountain terraced field type; and Water and Field Pattern (S6), which includes Field-shaped, Straight strip-shaped, and Free-form categories. The V group also consists of six subcategories: Shape Index of Settlement Boundary (V1), classified as Finger-shaped, Clustered, Cluster-belt-shaped, and Belt-shaped; Degree of Fragmentation of Settlement Boundary (V2), which includes Simple and smooth, and Complex and uneven; Degree of Dispersion of Village Plots (V3), divided into Dispersed system and Aggregated system; Colony Density (V4), categorized as Low-density parcels and High-density parcels; and Mean Area of Cluster Plots (V5), including Large-parcel settlements and Small-parcel settlements. The R group contains six subcategories as well: Relationship between Road Network Alignment and Landscape (R1), classified as Road network along water and mountains, and Road network crossing water systems; Relationship between Road Network Alignment and Wind Direction (R2), including Main road facing wind, Main road avoiding wind, and Low wind-direction relevance; Road Network Curvature (R3), divided into Low curvature, Medium curvature, and High curvature; Density of Road Network (R4), categorized as High road network density and Low road network density; Average Width of Road Network at Various Levels (R5), including Narrow road network settlements and Wide road network settlements; and Road Network Structure (R6), classified as Grid pattern, Fishbone pattern, and Mixed pattern. This classification standard is based on Duan Jin’s theory of spatial genes \cite{Duan2019SpaceGene,Cheng2022SpatialGeneMap}. Each subtype corresponds to distinct different spatial category, represented by numbers of class (e.g., cls-4 means that there are 4 classes for a subtype). Detail information are shown in Table \ref{independent}.

\begin{figure*}[t]
\centering
\includegraphics[width=1\linewidth]{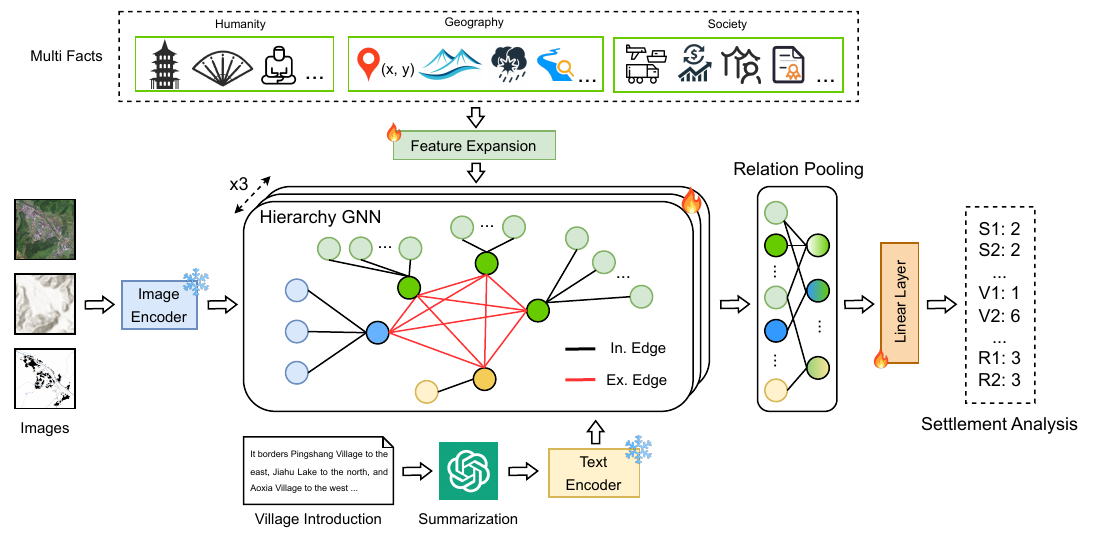}
\caption{Pipeline of proposed method. It uses a frozen image encoder and a text encoder to extract features from images and text. A learnable feature expansion module aligns the dimensions of multi-fact data with those of the image and text. A Graph Neural Network (GNN) propagates information across data types in a hierarchical structure. Finally, a relation pooling operation is applied, followed by a linear layer to predict all spatial morphology subtypes.}
\label{fig:overview}
\end{figure*}

\subsection{Task Definition}
Our task is to implement a comprehensive spatial morphology analysis of a village using multi-source information. It is a rule-based classification task for all 17 subtypes. The classification has to be guided by well-established principles, ensuring that the integration of diverse data sources adheres to recognized frameworks and methodologies. As shown in Figure \ref{fig:relation}, we show the principles of related multi-facts data for a subtype prediction. Note that, image and text data are not associated with any subtype. By leveraging these principles, we aim to provide a nuanced understanding of the village's spatial structure, incorporating both traditional and contemporary data sources for a more holistic analysis. It is quite challenging, as the integration of diverse data sources requires careful consideration of their varying data types, formats, and contextual relevance. Additionally, aligning these data sources with established principles for all spatial morphology sub-types necessitates meticulous understanding and harmonization processes. We thus designed a hierarchy GNNs to address this challenging task.

\subsection{Analysis Pipeline}

In Figure \ref{fig:overview}, we present an overview of our analysis pipeline. It employs a frozen image encoder and a text encoder to extract image and text features, respectively. A learnable feature expansion module is then applied to align the feature dimensions of multi-facts data with those of the image and text. After the feature extraction process, a Graph Convolutional Network (GCN) is utilized to propagate information across various data types within a hierarchical structure. Finally, a relation pooling operation is applied based on established principles, and a linear layer is used to forecast each spatial morphology sub-types of an input village.

We utilize pre-trained Contrastive Language-Image Pre-training (CLIP) \cite{radford2021learning} for the feature extraction purpose. CLIP, developed by OpenAI, is a model designed to learn multimodal representations through contrastive learning on a vast dataset of 400 million image-text pairs. It embeds both images and texts into a shared vector space, optimizing the alignment between matching pairs to maximize similarity. This enables CLIP to function as a robust backbone for image and text feature extraction, converting them into numerical vector representations that can be leveraged for a variety of downstream tasks. Let all image data be denoted as $\{x_i\}$, where $i$ represents the index of each image. A frozen CLIP image encoder will then extract the image features as follows:
\begin{equation}
    f^x_i = CLIP_{img}(x_i).
\end{equation}
Thus, all the image features can be represented as $\{f^x_i\}$. 

In a separate branch, we encode the text data (both text data and multi-facts with text) using CLIP's text encoder. However, the original text, which provides a brief introduction to a village, exceeds the token length limit of CLIP’s encoder (less than 77 tokens). To address this, we employ GPT-4o to summarize (We manually check the quality of summarization and a discussion for this operation is available in Section \ref{dis}) the introduction with the prompt: "\textit{Summarize the following text into a brief version with no more than 150 words}." Additionally, we utilize a longer version of the frozen CLIP text encoder with a maximum input length of 248 tokens for embedding the text features. Let $t$ represent the text input; the feature extraction process can be formulated as follows:
\begin{equation}
    f^t = CLIP_{text}(sum(t)),
\end{equation}
where $sum$ is the summarization using GPT-4o.

For multi-fact information with scalar, a learnable feature expansion layer is employed to achieve dimension alignment. Let the collection of facts be denoted as $\{r_{pq}\}$, where $p$ represents the index of the fact group, and $q$ indicates the index of a fact within group $p$. The feature expansion layer consists of a single fully connected layer that transforms a scalar input into a vector with the same dimensionality as the image and text features (e.g., 512 in the case of the CLIP model). The expansion process is formulated as follows:
\begin{equation}
    f^r_{pq} = Expansion(r_{pq}).
\end{equation}
Consequently, all the multi-fact features can be represented as $\{f^r_{pq}\}$. 

Leveraging the characteristics of multi-source information, we design a hierarchical graph structure. There are two types of nodes: \textbf{Input Nodes}, represented by light-colored nodes, which are derived from feature vectors of multi-source data. Image features are represented by light blue, multi-fact features are represented by light green, and text feature is represented by light yellow. We also designed \textbf{Communication Nodes}, depicted as dark-colored nodes, which are initialized randomly. The communication node is designed to facilitate information transfer between different modalities within a hierarchical structure. In our hierarchical graph, input nodes carry explicit geographic semantics. Image nodes represent spatial data such as satellite maps and topographic images; multi-fact nodes encode geospatial indicators; and text nodes provide descriptive information tied to specific villages. These nodes reflect real-world spatial or contextual elements. In contrast, communication nodes are abstract, non-geographic constructs designed to facilitate feature exchange across modalities. They do not correspond to physical entities but serve as intermediaries to enhance information propagation within the graph. Our approach leverages GNN's information propagation capabilities and multi-source data's structural properties to ensure orderly and efficient information flow. Unlike fully connected or Transformer-based architectures, our design prioritizes preserving data characteristics to the greatest extent.

In our design, the number of input nodes is strictly determined by the number of data modalities available for each village unit (e.g., spatial layout, land use, and environmental factors). Similarly, the communication nodes are constructed based on the major functional categories defined in the dataset (e.g., Humanity, Geography). This ensures that both types of nodes are directly grounded in the structure and semantics of the data, rather than arbitrarily set. Therefore, we did not conduct additional experiments by varying the number of communication nodes, as this would undermine the semantic consistency of the graph and deviate from the real-world data structure we aim to model.

Let's denote $G = \{g_1, g_2, \ldots, g_N\}$ for light-colored nodes (feature vectors of multi-source input), each of which has a dimension of $d$ and $H = \{h_1, h_2, \ldots, h_M\}$ for dark-colored nodes (randomly initialized communication nodes), each of which has a dimension of $d$. For the initial layer of $l=0$, multi-source node features directly use the input feature vector $G$ and communication nodes is initialized as:
\begin{equation}
    H^{(0)} = \text{Random}([M, d]).
\end{equation}

There are also two kinds of edges. \textbf{Input Edges} (black edges) connect the input nodes $G$ and the communication node $H$. Following the definition of GCN, the adjacent matrix is represented as $A_{in} \in \mathbb{R}^{N \times M}$, where $A_{in}[i,j] = 1$ means there is connection between $g_i$ and $h_j$. Note that the connection between $G$ and $H$ is predefined by our data structure. The normalized adjacency matrix can be formulated as:
\begin{equation}
    \tilde{A}_{in} = D_{in}^{-\frac{1}{2}} A_{in} D_{in}^{-\frac{1}{2}}
\end{equation}
where $D_{in}$ is degree matrices. \textbf{Communication Edges} (red edges) connect communication nodes $H$. Instead of the fixed adjacent matrix, we applied a GAT for dynamic calculation. 

A hierarchical GNN for feature propagation and updating is also proposed. It is started from GCN feature propagation between input nodes and communication nodes as follows:
\begin{equation}
    H_{in}^{(l+1)}, G^{(l+1)} = \sigma\left(\tilde{A}_{in} [G^{(l)}H^{(l)}] W_{in}^{(l)}\right),
\end{equation}
where $W_{in}^{(l)} \in \mathbb{R}^{d \times d}$ is a transformation weight matrix between input features and communication nodes. $\sigma(\cdot)$ is a non-linear activation function (such as ReLU).

Then, feature fusion between communication nodes is implemented using GAT as:
\begin{equation}
    e_{ij} = \text{LeakyReLU}\left(\mathbf{a}^\top [\mathbf{h}^{l+1}_i || \mathbf{h}^{l+1}_j]\right)
\end{equation}
where $e_{ij}$ is edge attention score between two node in $H_{in}^{(l+1)}$. $\mathbf{h}^{l+1}_i$ and $\mathbf{h}^{l+1}_j$ is the input feature of node $i$ and $j$. $\mathbf{a}$ is a learnable attention weight vector and $||$ represents the concatenation operation of features. Then, we normalize the edge attention weights via the following function:
\begin{equation}
    \alpha_{ij} = \frac{\exp(e_{ij})}{\sum_{j \in \mathcal{N}(i)} \exp(e_{ij})},
\end{equation}
where $\mathcal{N}(i)$ is the set of neighbors of node $i$ and $\alpha_{ij}$ is the attention weight of node j to node i. The next layer of features of node $i$ is calculated by weighted aggregation of neighboring node features as follows:
\begin{equation}
    \mathbf{h'}_i^{(l+1)} = \sigma\left(\sum_{j \in \mathcal{N}(i)} \alpha_{ij} \mathbf{W^{(l)}_{ex}} \mathbf{h}_j^{(l+1)}\right),
\end{equation}
where $\mathbf{h'}_i^{(l+1)}$ is the output feature of node $i$ at layer $l+1$, $\mathbf{W_{ex}}$ is a learnable weight matrix, and $\sigma$ is the Relu activation. The combination of all updated nodes at layer $l+1$ can be represented by $H_{ex}^{(l+1)}$ and the feature update of the communication node is based on $H_{in}^{(l+1)}$ and $H_{ex}^{(l+1)}$, expressed as:
\begin{equation}
    H^{(l+1)} = \beta H_{in}^{(l+1)} + (1 - \beta) H_{ex}^{(l+1)},
\end{equation}
Where $\beta \in [0, 1]$ is the fusion weight coefficient (default as 0.6). Let's represent the above propagation process as HGNN and after the $L$ layers of graph operations (we implement 3 times), the final features of the communication nodes and input nodes are:
\begin{equation}
    H^{(L)}, G^{(L)} = \text{HGNN}(G^{(0)}, H^{(0)}, A_{in}).
\end{equation}
$G^{(L)}$ is used as a high-dimensional feature representation for the subsequent relation pooling and classification modules.

\begin{figure*}[t]
\centering
\includegraphics[width=1\linewidth]{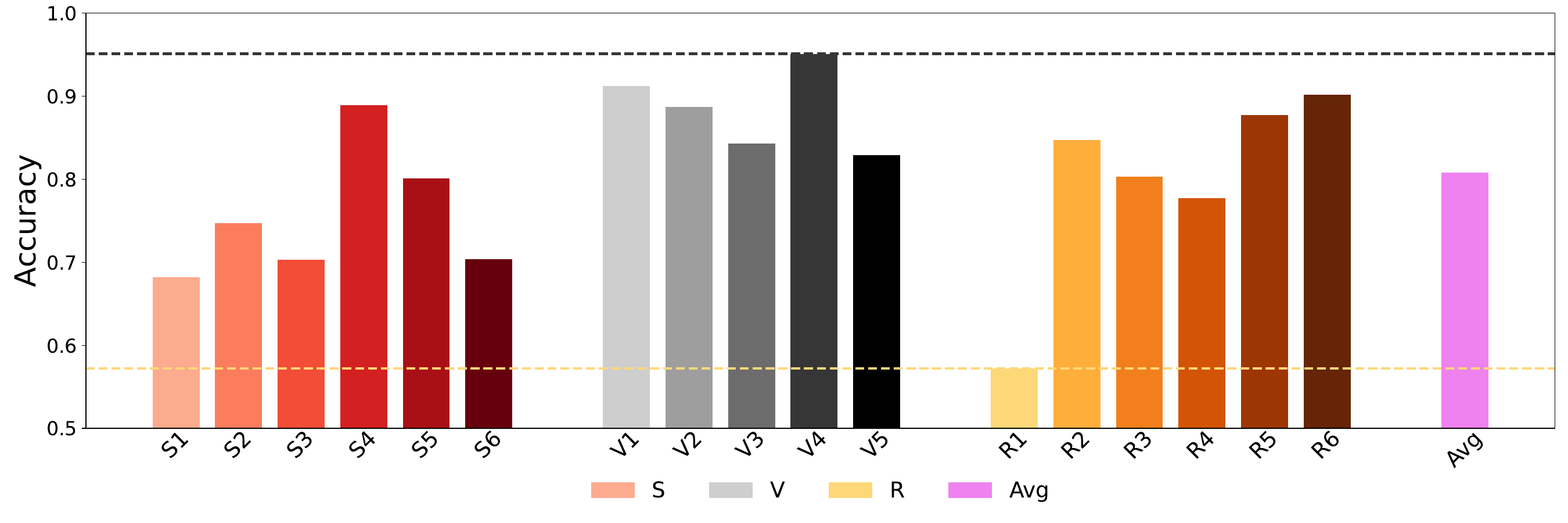}
\caption{Overall accuracy across all subtypes of village spatial morphology is presented. The results for S types, V types, and R types are represented in gradient red, grey, and gold, respectively. The average (Avg) accuracy for all subtypes is depicted in light purple. Additionally, dotted lines indicate the best and worst predicted types, respectively.}
\label{fig:over_acc}
\end{figure*}

To ensure strict adherence to the established principles for the classification of all subtypes, we developed a relation pooling mechanism. In this design, each subtype leverages the fusion of its corresponding input sources following the processing by the HGNN. Additionally, we implemented a joint training strategy that facilitates simultaneous optimization for all subtype classifications. This collaboration promotes the sharing of learned representations across subtypes, thereby improving the overall model performance and the efficiency in prediction.

For a given subtype $k \in K$, we apply a direct averaging of the associated input source features ($G^{(L)}$) after processing through the final layer of the HGNN framework. The association between the subtype and the input sources is defined in Section Dataset Construction. We denote this mapping for subtype $k$ as $P^k$, which specifies the indices of the corresponding input sources. This approach enables subtype-specific feature aggregation based on predefined relations. The pooling operation for subtype $k$ can thus be mathematically expressed as follows:
\begin{equation}
    f^k = avg(G^{(L)} \vee P^k),
\end{equation}
where $avg$ is the average operation and $\vee$ is the feature selection. Continuously, for each subtype $k$, we use a single full connection layer independently as a classification head to predict the final subtype. We also calculate the corresponding classification loss (cross entropy loss) for each classification head separately, and then weighted sum these losses to get the total loss function. Specifically, assuming that the prediction of the \(k\)th classification head with feature $f^k$ is $\hat{y}_k$, the corresponding true label is $y_k$, and its loss function is $\mathcal{L}_i(\hat{y}_k, y_k)$. The total loss function can be defined as:
\begin{equation}
    \mathcal{L}_{\text{total}} = \frac{1}{K} \sum_{k=1}^{K} \mathcal{L}_k(\hat{y}_k, y_k).
\end{equation}
Our experimental results demonstrate that this joint training approach mitigates overfitting caused by insufficient data, as discussed in Section \ref{ablation}.

\section{Results}
\subsection{Experimental Settings}
We divided the dataset into three parts: 80\% for training and 20\% for testing. During training, we saved the model checkpoint in the final epoch. Subsequently, all reported results are based on the evaluation of the saved model on the test set. For our hierarchical GNN model, training was conducted for a total of 20 epochs using a learning rate of 0.0001 and the AdamW optimizer. All experiments were implemented on a GPU server equipped with 4 NVIDIA A40 GPUs. 

\begin{table*}[t]
\caption{Performance comparison of our method to recent state-of-the-art fusion methods. For the parameters calculation, we use $M = 10^6, B = 10^9$ as the statistical unit.}
\label{comparision}
\centering
\resizebox{0.8\textwidth}{!}{
\begin{tabular}{lccccccccc}
\toprule
& Params & \multicolumn{2}{c}{S} &\multicolumn{2}{c}{V} &\multicolumn{2}{c}{R} & \multicolumn{2}{c}{Average}\\
\cmidrule(lr){3-4} \cmidrule(lr){5-6} \cmidrule(lr){7-8} \cmidrule(lr){9-10}
Methods & (M/B) & Acc  & F1 & Acc  & F1 & Acc  & F1 & Acc  & F1 \\ 
\midrule
Full Connection & 86.2M & 0.598 & 0.682 & 0.650 & 0.713 & 0.627 & 0.690 & 0.626 & 0.684\\
Transformer & 92M & 0.703 & 0.790 & 0.760 & 0.864 & 0.801 & 0.887 & 0.755 & 0.867\\
\midrule
GCN \cite{kipf2016semi} & 87.1M & 0.707 & 0.796 & 0.809 & 0.897 & 0.800 & 0.891 & 0.772 & 0.861 \\
GAT \cite{velickovic2017graph} & 87.3M & 0.712 & 0.807 & 0.816 & 0.902 & 0.806 & 0.895 & 0.778 & 0.866 \\
Graphormer \cite{ying2021transformers} & 91M & 0.703 & 0.804 & 0.851 & 0.921 & 0.839 & 0.912 & 0.794 & 0.874 \\
\midrule
MBT \cite{Nagrani2021} & 143M & 0.702 & 0.816 & 0.813 & 0.903 & 0.817 & 0.898 & 0.752 & 0.857\\
DD-IMvMLC-net \cite{Wen2023} & 51M & 0.517 & 0.580 & 0.580 & 0.684 & 0.591 & 0.660 & 0.562 & 0.643\\
DDAG \cite{Ye2020} & 28M & 0.587 & 0.672 & 0.630 & 0.692 & 0.647 & 0.710 & 0.621 & 0.680\\
VLC-BERT \cite{Ravi2023} & 110M & 0.710 & 0.800 & 0.812 & 0.900 & 0.803 & 0.894 & 0.775 & 0.868 \\
X2L \cite{chen2023x} & 6B & 0.725 & 0.837 & 0.845 & 0.920 & 0.831 & 0.907 & 0.793 & 0.889\\
LLaVA‑NeXT‑7B \cite{li2024llava} & 7B & 0.710 & 0.820 & 0.848 & 0.918 & 0.837 & 0.910 & 0.791 & 0.880\\
Qwen2.5‑VL‑7B \cite{bai2025qwen2} & 7B & \textbf{0.735} & \textbf{0.844} & 0.860 & 0.926 & 0.835 & 0.911 & 0.806 & 0.892\\
\midrule
\textit{Ours} & 93M & 0.720 & 0.833 & \textbf{0.884} & \textbf{0.933} & \textbf{0.842} & \textbf{0.919} & \textbf{0.821} & \textbf{0.907} \\
\bottomrule
\end{tabular}}
\end{table*}

\subsection{Prediction Accuracy of Proposed Method}
As illustrated in Figure \ref{fig:over_acc}, we present a detailed breakdown of task-wise accuracy, represented by distinct color-coded bars: red variants for S, grey variants for V, and golden variants for R. Additionally, the light purple bar depicts the average performance across all tasks, providing a clear visual summary of the overall trends and comparative accuracy for each category. For settlement landscape spatial structure (S1-S6), the accuracies vary significantly, with S4 achieving the highest accuracy, exceeding 0.9, while S1 has the lowest accuracy, below 0.7, indicating notable differences in classification performance across tasks. Compared with other groups, the overall classification accuracy is lower. For settlement and plot morphological patterns (V1-V5), the accuracies are relatively consistent, with V4 reaching nearly 0.95 and others are slightly lower but still above 0.8, demonstrating stable performance for this task group. For settlement road network patterns (R1-R6), the accuracies show greater variation. R5 and R6 achieve high accuracies above 0.9, whereas R1 has the lowest accuracy, just above 0.6, suggesting significant differences in task complexity and model performance requirements. The overall average accuracy (Avg) is approximately 0.85, reflecting good overall performance but with room for improvement, particularly for tasks with lower accuracies, such as S6 and R1. These variations in accuracy may indicate differences in task complexity or data quality and distribution, suggesting that future research could focus on optimizing feature selection or model design for low-performing tasks to further enhance overall performance.

We also implement GCN, GAT, and Graphormer \cite{ying2021transformers} for graph-based structure comparison. Among them, Graphormer is a Transformer-based model that incorporates graph structural information by encoding node positions, edge relations, and centrality features. We leverage its structured attention mechanism to perform unified modeling and fusion of multi-source inputs. To compare our approach with other multi-source fusion technologies, we also reproduced four recent popular methods in this research direction \cite{Gao2020}. Among them, DD-IMvMLC-net \cite{Wen2023} introduced a self-supervised pretraining method for multi-view input. This method employs an encoder-fusion-decoder architecture for representation learning, enabling the learned encoder to support various downstream tasks, including classification. We adopted its structure for our multitask pretraining, followed by classification. DDAG \cite{Ye2020} utilized a graph-based structure for information fusion in person re-identification. In our task, their cross-modality graph-structured attention approach is adopted. VLC-BERT \cite{Ravi2023} employs a BERT model to integrate multiple input sources. While similar to other transformer-based methods, it distinguishes itself with a larger model size. X2L \cite{chen2023x} leverages recent advancements in Large Language Models (LLMs) for information fusion. A frozen LLM produces responses by using multi-source input embeddings generated by corresponding frozen encoders. Similarly, we adopt this design to utilize LLMs for class prediction by fine-tuning a classification head on top of the LLM for each subtype and LORA fine-tuning for LLMs. In response to recent advances in multimodal large language models, we further evaluated our task using two representative models: LLaVA-NeXT-7B \cite{li2024llava} and Qwen2.5-VL-7B \cite{bai2025qwen2}. Following the same X2L-like setup, we froze both the visual encoder and the LLM decoder, and fine-tuned a lightweight classification head for category prediction. We also compared our method with two basic types of multi-source fusion methods: fully connected fusion and transformer-based fusion. For the former, we applied a fully connected layer directly after relation pooling without incorporating our HGNN. For the latter, we used a transformer model for fusion, leveraging a global attention mechanism.

To ensure fairness and reproducibility in comparative and ablation studies, we provide detailed descriptions of the model architectures used in our experiments. For the visual encoder, we adopt the official CLIP ViT-B/32 model released by OpenAI for all graph-based methods. The transformer for transformer-based fusion consisting of 2 layers, each with a hidden size of 256, 4 attention heads, and a feed-forward dimension of 512. Additionally, we compute the total number of parameters (including encoder) for all compared methods to ensure a fair comparison in terms of model complexity. The results are reported in Table 2. Note that all results for existing methods were reproduced on our dataset using their source code with joint training. However, there may be some errors or inconsistencies in the reproduction process.

In Table \ref{comparision}, we present a comparison of results between our method and other techniques. The results demonstrate that our hierarchical GNN achieves the best performance in all tasks, significantly outperforming the basic fully connected approach. This indicates that merely merging features is insufficient for improving predictive accuracy. While transformer-based fusion performs better than full connections, indiscriminately attending to all sources does not appear to benefit all tasks effectively. Traditional GCN and GAT models already outperform non-graph baselines such as Full Connection and Transformer, indicating the advantage of incorporating structural relationships in multimodal fusion. Graphormer, which introduces Transformer-like global attention into the graph domain, further improves the performance over GCN and GAT, particularly on the V and R subsets. Notably, our proposed HGNN achieves the best performance across all three tasks (S, V, R), surpassing Graphormer by a obvious margin in terms of both accuracy and F1 score. This highlights the effectiveness of our hierarchical structure in capturing the multi-level spatial and semantic dependencies among multimodal features. These results demonstrate that careful modeling of graph topology and feature hierarchy plays a crucial role in enhancing fusion performance, beyond merely applying standard graph models. It emphasizes the importance of using a predefined graph structure for information fusion. For other fusion methods, VLC-BERT demonstrates performance comparable to that of the transformer, indicating that increasing the model size does not necessarily improve results. DD-IMvMLC-net performs poorly, suggesting that its pretraining approach is unsuitable for our task. We believe that the limited data quantity negatively impacts the original CLIP representation. Similarly, DDAG exhibits low performance, highlighting that its undefined graph structure fails to enhance the model’s effectiveness. One particularly intriguing result is that the LLMs-based method X2L demonstrates exceptional performance, highlighting the significant potential of leveraging Large Language Models for multi-source data fusion. We can also find that LLaVA-NeXT-7B achieved comparable performance to X2L, while Qwen-7B performed slightly better, reflecting the enhanced alignment and reasoning capacity of more recent LLMs. However, we observe that these general-purpose models still lack the ability to explicitly model the spatial structure and hierarchical relations inherent in traditional village morphology data. In contrast, our HGNN approach effectively incorporates such structural priors through hierarchical graph construction and multimodal feature fusion, resulting in superior interpretability and robustness. These findings demonstrate that while LLMs offer strong generalization, dedicated structural modeling—as in HGNN—remains crucial for tasks involving complex spatial semantics. Future research could further explore this capability, optimizing LLMs for even more complex fusion tasks across various domains.

\begin{figure}[t]
\centering
\includegraphics[width=1\columnwidth]{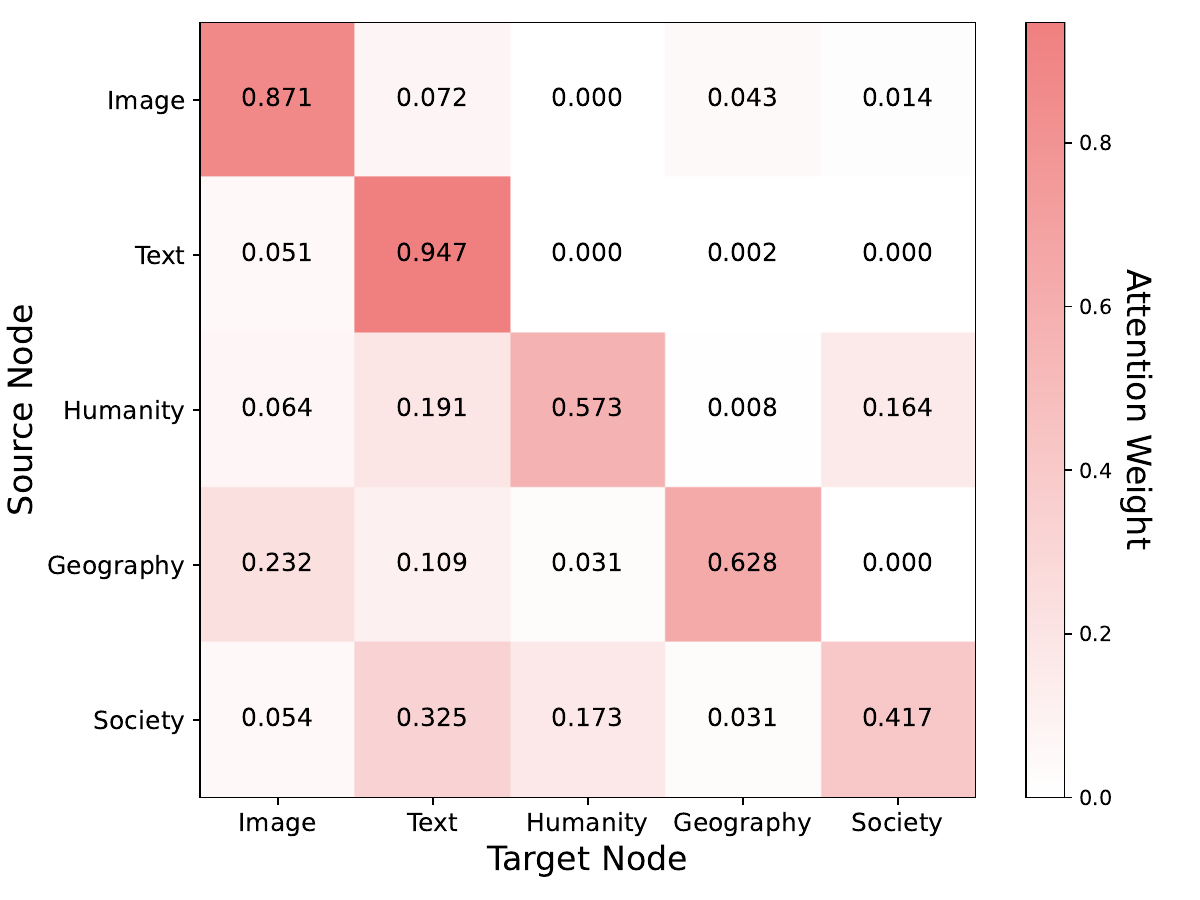}
\caption{Relation strength among all communication nodes.}
\label{fig:graph}
\end{figure}

\subsection{Relation Analysis}
The advantage of using GAT is the ability to directly visualize the weight between nodes, offering insights into their relationships. We visualized the attention weights between nodes in the final layer of the graph attention network, as shown in Figure \ref{fig:graph}. The connection between Image and itself has the highest weight (0.871), reflecting the strong importance of preserving node-specific features for Image. Similarly, Text shows a high self-attention weight (0.947), suggesting that textual features are primarily self-contained and less dependent on other nodes. Among inter-node relationships, Society and Text exhibit a notable connection with a weight of 0.325, indicating a moderate level of information exchange or semantic overlap. The relationship between Geography and Image (weight 0.232) hints at some dependence or shared features in the learning process.

In contrast, relationships such as Image to Humanity and Humanity to Image have weights close to zero (0.000 and 0.064, respectively), suggesting minimal interaction or influence between these nodes. Similarly, Geography exhibits relatively weak connections with Text (0.109) and Humanity (0.031), but maintains a strong self-attention weight (0.628), emphasizing the node's independence. Overall, the network highlights strong self-connections for Image and Text, with inter-node relationships concentrated primarily between Society, Text, and Geography. These insights suggest that the task relies on the features of nodes like Image and Text, while connections among Society, Geography, and Humanity play a more collaborative role in the learning process.

\begin{table}[t]
\centering
\small
\caption{Comparison of three different training strategies across three tasks.}
\label{compare_training}
\resizebox{1\linewidth}{!}{
\begin{tabular}{lcccccccc}
\toprule
\multirow{2}{*}{Methods} & \multicolumn{2}{c}{S} & \multicolumn{2}{c}{V} & \multicolumn{2}{c}{R} & \multicolumn{2}{c}{Average} \\
\cmidrule(lr){2-3} \cmidrule(lr){4-5} \cmidrule(lr){6-7} \cmidrule(lr){8-9}
& Acc & F1 & Acc & F1 & Acc & F1 & Acc & F1 \\
\midrule
Split Training   & 0.685 & 0.774 & 0.740 & 0.845 & 0.707 & 0.823 & 0.710 & 0.829 \\
Group Training   & 0.709 & 0.826 & 0.856 & 0.923 & 0.820 & 0.897 & 0.795 & 0.881 \\
Overall Training & \textbf{0.720} & \textbf{0.833} & \textbf{0.884} & \textbf{0.933} & \textbf{0.842} & \textbf{0.919} & \textbf{0.815} & \textbf{0.902} \\
\bottomrule
\end{tabular}}
\end{table}

\subsection{Ablation Study} \label{ablation}
Our approach integrates multiple data sources, diverse model configurations, and advanced learning strategies to enhance performance and robustness. In this section, we conduct a comprehensive analysis, delving into the impact of each component and examining how their interplay contributes to the method's overall effectiveness.

Our dataset consists of three main groups with a total of 17 sub-types. A common approach is to train a separate model for each sub-type; however, this often leads to overfitting due to data limitations and the oversimplification of tasks. To overcome these challenges, we adopted a joint training strategy that integrates all sub-type predictions into a unified framework. To identify the optimal strategy, we compared three training approaches: (1) Split Training, where each sub-type is trained independently; (2) Group Training, which involves joint training within each group of related sub-types; and (3) Overall Training, where all sub-types across groups are trained jointly in a single model. The results in Table \ref{compare_training} show that joint training with subtype classification improves the performance of spatial morphology classification, particularly on complex spatial forms such as hybrid patterns. However, we acknowledge that this improvement may not solely result from the model learning the underlying generation logic. It is also possible that correlations between frequently co-occurring subtypes in the dataset contribute to this gain. We leave a more rigorous investigation of the model's ability to capture structural semantics for future work.

\begin{figure}[t]
\centering
\includegraphics[width=1\columnwidth]{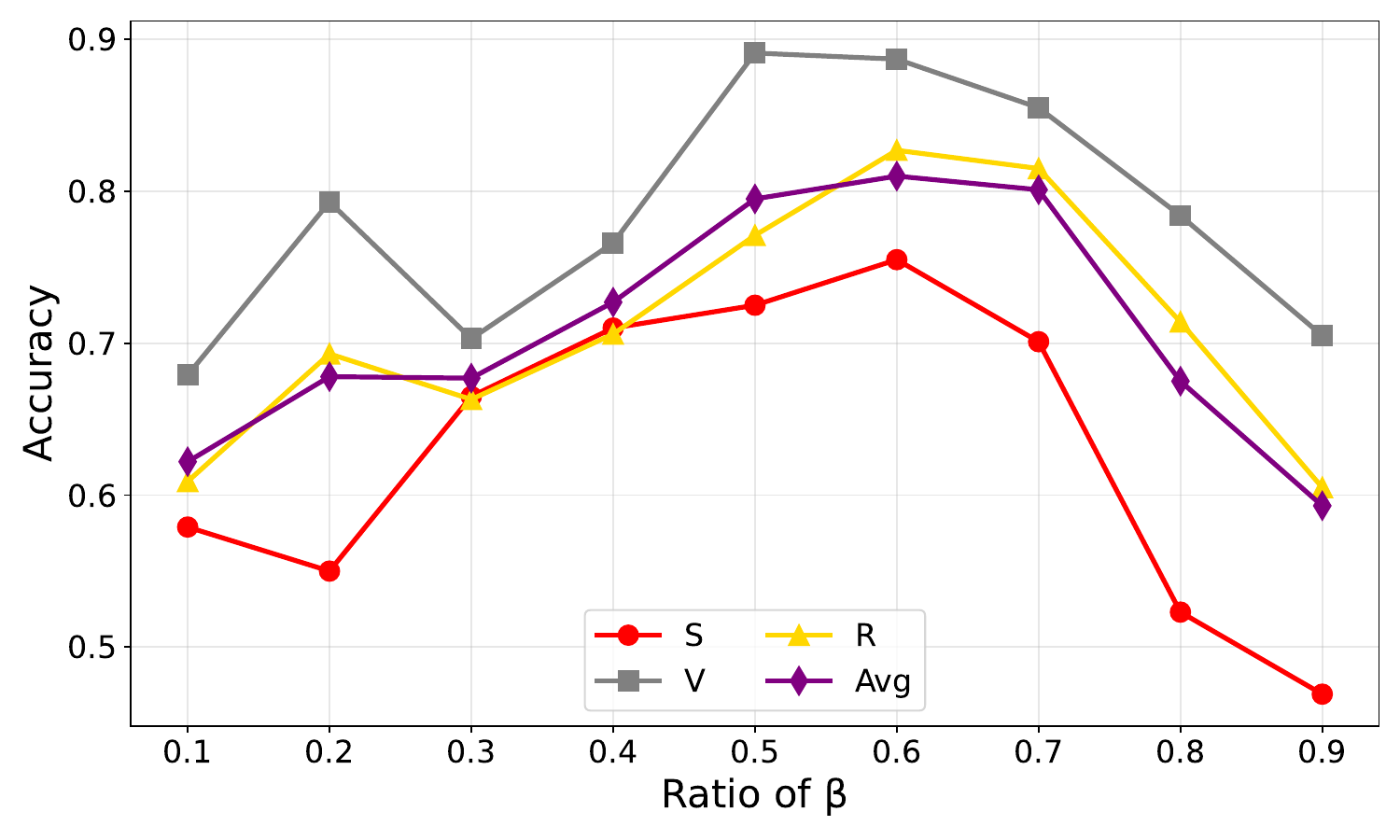}
\caption{Impact of using different $\beta$ value for communication nodes fusion.}
\label{fig:beta}
\end{figure}

The impact of weight coefficient $\beta$ for fusion of GCN and GAT is also important. Figure \ref{fig:beta} illustrates the impact of different $\beta$ coefficients on the accuracy of tasks S, V, and R, as well as the overall average accuracy (Avg). Overall, task S (red line) performs poorly at lower $\beta$ values (0.1-0.3), gradually improves as $\beta$ increases, reaches a peak around 0.6, and then declines sharply. Task V (gray line) exhibits a more stable trend, with accuracy steadily increasing as $\beta$ rises, peaking around 0.3, and then gradually decreasing. Task R (yellow line) starts with low accuracy at smaller $\beta$ values, improves significantly in the middle range, and declines at higher $\beta$ values. The overall average accuracy (purple line, Avg) reflects a balanced trend across tasks, showing optimal performance in the mid-range of $\beta$ values (0.3-0.6), while both excessively high and low $\beta$ values result in reduced accuracy. Further analysis reveals that task S is the most sensitive to changes in $\beta$, with large fluctuations in accuracy, whereas task V is less dependent on $\beta$, exhibiting a more stable trend, and task R shows moderate fluctuations.

Overall, $\beta$ values between 0.3 and 0.6 yield the highest average accuracy, making this range a critical region for balancing model performance. Lower $\beta$ values may lead to insufficient weighting of certain features, while higher values may result in over-optimization for specific tasks. Further optimizing model parameters within the mid-range of $\beta$ values or adjusting $\beta$ based on task priorities may enhance the performance and we take this as our future research.

\begin{figure}[t]
\centering
\includegraphics[width=1\columnwidth]{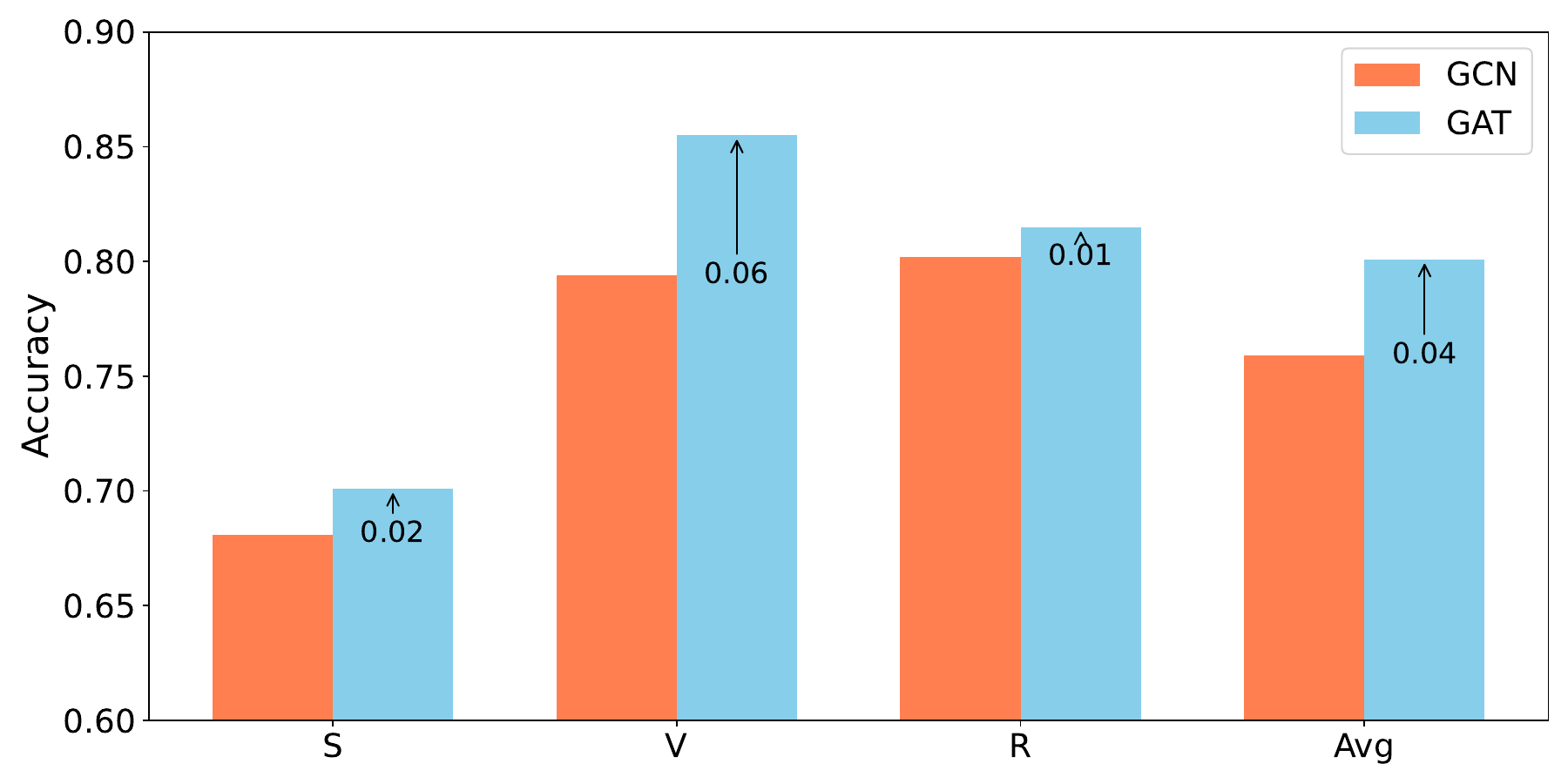}
\caption{Advantage of using GAT as communication edges.}
\label{fig:GAT}
\end{figure}

\begin{table}[t]
\centering
\small
\caption{Comparison of different initialization strategies for communication nodes.}
\label{compare_init}
\resizebox{1\linewidth}{!}{
\begin{tabular}{lcccccccc}
\toprule
\multirow{2}{*}{Initialization Strategy} & \multicolumn{2}{c}{S} & \multicolumn{2}{c}{V} & \multicolumn{2}{c}{R} & \multicolumn{2}{c}{Average} \\
\cmidrule(lr){2-3} \cmidrule(lr){4-5} \cmidrule(lr){6-7} \cmidrule(lr){8-9}
& Acc & F1 & Acc & F1 & Acc & F1 & Acc & F1 \\
\midrule
Random Init (Default)  & 0.720 & 0.833 & 0.884 & 0.933 & 0.842 & 0.919 & 0.815 & 0.902 \\
Mean Feature Init           & 0.717 & 0.828 & 0.871 & 0.922 & 0.834 & 0.908 & 0.807 & 0.896 \\
\bottomrule
\end{tabular}}
\end{table}

\begin{table}[ht]
\centering
\small
\caption{Feature alignment via an FC layer.}
\label{compare_fc}
\resizebox{1\linewidth}{!}{
\begin{tabular}{lcccccccc}
\toprule
\multirow{2}{*}{Setting} & \multicolumn{2}{c}{S} & \multicolumn{2}{c}{V} & \multicolumn{2}{c}{R} & \multicolumn{2}{c}{Average} \\
\cmidrule(lr){2-3} \cmidrule(lr){4-5} \cmidrule(lr){6-7} \cmidrule(lr){8-9}
& Acc & F1 & Acc & F1 & Acc & F1 & Acc & F1 \\
\midrule
w. FC   & 0.720 & 0.833 & 0.884 & 0.933 & 0.842 & 0.919 & 0.815 & 0.902 \\
w.o. FC & 0.681 & 0.802 & 0.853 & 0.912 & 0.820 & 0.907 & 0.788 & 0.870 \\
\bottomrule
\end{tabular}}
\end{table}

Effective communication between multi-source information is crucial for information fusion and subsequent classification of subtypes. To achieve this, the communication node is designed to facilitate information transfer across different modalities within a hierarchical structure. We employed GAT for dynamic relation learning, but a fixed relation approach using a simple GCN can also enable communication. Therefore, we compare the results of GCN and GAT under the same experimental settings. As shown in Figure \ref{fig:GAT}, the performance differences between GAT and GCN are illustrated across all three groups and their average performance.

\begin{table*}[ht]
  \caption{A demonstration of village introduction compression through GPT-4o.}
  \centering
  \begin{tabular}{>{\raggedright\arraybackslash}m{4cm} >{\raggedright\arraybackslash}m{12cm}}
    \toprule
    \textbf{Raw Introduction} & Zhenggui Village, Liren Town, is located in the southwest of Liren Town, 8 kilometers away from Longnan County, 2 kilometers away from Liren Town, and 1 kilometer away from National Highway 105. Liren Town has convenient transportation and significant location advantages. National Highway 105 spans from east to west, Beijing-Kowloon Railway and Gan-Yue Expressway run from north to south, and Gan-Shen High-speed Railway passes through the town. It is the only intersection of the three major arteries of national highways, expressways, and railways in Longnan County, and it is also an important transportation hub in Longnan County. Zhenggui Village has an area of 1.97 square kilometers, 900 mu of cultivated land, and 1,400 mu of forest land. With lush ancient trees and a pleasant environment, it is an ancient Hakka village with a history of more than 500 years. \\
    \midrule
    \textbf{GPT-4o Compression} & Zhenggui Village, in Liren Town, lies 8 km from Longnan County and 1 km from National Highway 105. Liren Town is a key transportation hub, intersected by National Highway 105, the Beijing-Kowloon Railway, Gan-Yue Expressway, and Gan-Shen High-speed Railway. Zhenggui Village spans 1.97 square kilometers, with 900 mu of cultivated land and 1,400 mu of forest. Known for its lush ancient trees and pleasant environment, it is an ancient Hakka village with over 500 years of history. \\
    \bottomrule
  \end{tabular}
  \label{table_sample_gpt4}
\end{table*}

It is obvious that GAT outperforms GCN across all groups (S, V, R) and in average performance (Avg). Specifically, GAT achieves 2\% higher accuracy than GCN in group S, with the most significant improvement observed in group V, where the accuracy is 6\% higher. In group R, the performance difference is minimal, with GAT only surpassing GCN by 1\% . Overall, GAT demonstrates a 4\% improvement in average performance compared to GCN. This indicates that GAT's dynamic relation learning is particularly effective in scenarios with more complex information interactions (e.g., group V), while its advantage diminishes in simpler scenarios (e.g., group R). In general, GAT shows clear advantages in enhancing multimodal information fusion and classification performance.

To evaluate the impact of initialization strategies for communication nodes, we conducted an ablation study comparing two methods: (1) random initialization, which serves as our default strategy, and (2) mean feature initialization, where the communication node is initialized with the average feature vector of all input nodes. The results are summarized in Table \ref{compare_init}. As shown, the performance difference between the two strategies is minor across all tasks. For example, the average accuracy and F1 score are 0.815 and 0.902 for random initialization, and 0.807 and 0.896 for mean feature initialization, respectively. This indicates that the initialization method has limited influence on the overall performance. The model is able to adapt and fuse features effectively regardless of the initial values, thanks to the hierarchical message-passing architecture and iterative fusion process. These findings demonstrate the robustness of our model to variations in communication node initialization.

To enhance the compatibility of heterogeneous features derived from different modalities, we incorporate an FC layer as a unified transformation module for feature expansion and alignment. Given that the feature distributions and dimensionalities vary significantly across modalities, directly aggregating or fusing them may result in suboptimal representations and hinder downstream reasoning. This transformation allows the model to capture higher-order relationships and complex patterns across modalities during the message passing process in the hierarchical graph structure. To verify the effectiveness of this component, we conducted an ablation study by removing the FC layer. As reported in Table~\ref{compare_fc}, the performance noticeably degrades without this transformation layer, with the average accuracy and F1 score dropping by 2.7\% and 3.2\%, respectively. These results confirm the necessity of the FC layer in ensuring effective multi-modal feature integration and maintaining the overall performance of our framework.

\section{Discussion} \label{dis}
This section provides a detailed discussion of the implementation specifics, current limitations, practical applications, and potential directions for future development.

As outlined in the methodology section, we employed GPT-4o to perform text compression for village introductions. This process aimed to produce concise summaries while retaining the essential information. Table \ref{table_sample_gpt4} illustrates this approach, presenting a sample village introduction in its original form alongside its compressed counterpart generated by GPT-4o. The prompt used for this task was: "Summarize the following text into a brief version with no more than 150 words." The outputs demonstrate that GPT-4o successfully preserves the key information from the original text while presenting it in a more succinct format. This capability ensures that the resulting summaries are both accurate and efficient for further analysis or presentation.

To evaluate whether the sentence compression by GPT-4o introduces semantic bias or leads to feature loss in rural scenarios, we conducted a detailed analysis (Table \ref{tab:compression-eval}) on all 118 village introductions from the test set. Two experts in architectural heritage from Nanchang University were invited to manually generate concise versions of each introduction (limited to 150 characters) and annotate the presence of three categories of key information: geospatial context, cultural symbols, and rural landscape. We then compared these expert-written summaries with the outputs generated by GPT-4o. For each sample, we manually assessed whether the critical information in each category was successfully preserved. In Table, The results showed that GPT-4o achieved high retention rates: with an overall average retention rate of 0.925 across all categories. In addition to manual evaluation, we used ROUGE-L and BERTScore to quantitatively assess the similarity between GPT-4o outputs and expert summaries. GPT-4o attained a ROUGE-L score of 0.614 and a BERTScore F1 of 0.822, indicating that the compressed outputs preserved not only key content but also structural and semantic coherence. These findings confirm that GPT-4o can effectively compress rural descriptive texts while retaining essential information with minimal semantic loss.

\begin{table}[t]
\centering
\caption{Evaluation of Introduction Compression by GPT-4o.}
\label{tab:compression-eval}
\resizebox{0.65\linewidth}{!}{
\begin{tabular}{lcc}
\toprule
\textbf{Method} & \textbf{Metric} & \textbf{Score} \\
\midrule
Manual & Retention Rate & 0.925 \\
\midrule
Automatic & ROUGE-L & 0.614 \\
          & BERTScore (F1) & 0.822 \\
\bottomrule
\end{tabular}
}
\end{table}

Our method leverages the characteristics of multimodal inputs and designs a hierarchical graph neural network (HGNN) that introduces two types of nodes—input nodes and communication nodes—as well as two types of edges—static input edges and dynamic communication edges. This design enables efficient multimodal feature fusion and propagation. By combining GCN and GAT in a two-stage feature update mechanism, the model not only preserves the unique properties of each modality but also enhances the flexibility of information exchange. Additionally, the relation pooling mechanism and joint training strategy improve subtype classification performance while effectively mitigating overfitting in small-sample datasets. The clear hierarchical structure and quantitative experimental results also enhance the interpretability and reliability of the proposed method.
    
Despite these advantages, the current framework still has several limitations. The random initialization of communication nodes may lead to instability in model performance. The static definition of input edges might overlook latent relationships among multimodal data. Furthermore, the dynamic edge computation and multi-stage feature update mechanism increase the model’s computational complexity, making it less scalable to large-scale datasets. The fixed fusion weight coefficient (0.6) lacks empirical or theoretical justification, and dynamically learned weights may provide better adaptability. We also explored the use of an adaptive fusion weight learning mechanism to further enhance model flexibility. However, we observed that the training process became significantly unstable under the current framework. As such, the adaptive strategy was not adopted in the final version of the model. We leave the integration of a more stable adaptive weighting approach for future work. Finally, the model is currently designed for fixed modality inputs and does not yet support variable modality numbers or dimensions.

\begin{figure}[t]
    \centering
    \includegraphics[width=1\columnwidth]{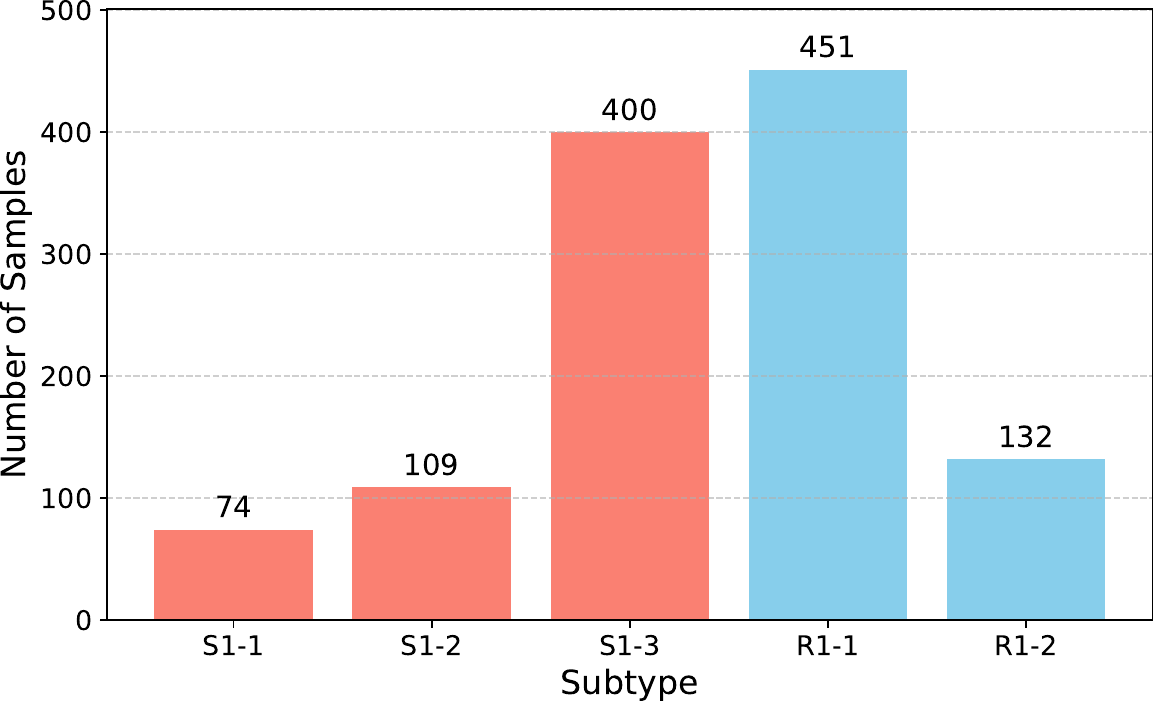}
    \caption{Data imbalance among classes for subtypes S1 and R1.}
    \label{fig:imblance}
    \end{figure}
    
We also observed that some subtypes, such as S1 and R1, show class imbalance, as shown in Figure \ref{fig:imblance}. To address this, we applied data augmentation (e.g., random flipping, rotation, and cropping) during training. We also tried using a class-weighted loss function by setting the weight parameter in the loss function. However, this did not improve the accuracy of the underrepresented subtypes. We will explore more effective methods for handling class imbalance.
    
In future work, we plan to address these issues from multiple perspectives. First, we aim to introduce prior-guided or data-dependent strategies for initializing communication nodes, thereby improving the model's stability and convergence. Second, we will explore adaptive edge construction mechanisms using similarity metrics, semantic proximity, or learnable attention to better capture complex cross-modal relationships. Third, we plan to incorporate lightweight GNN variants or sparsity-inducing techniques to improve computational efficiency and scalability on large-scale or high-resolution spatial datasets. Fourth, we intend to design a dynamic fusion module that learns optimal weights for each modality in an end-to-end manner, potentially leveraging reinforcement learning or contrastive learning for more robust multimodal alignment. Fifth, we will investigate architectural optimizations that enable the model to accommodate dynamic modality configurations, such as the real-time integration of streaming sensor data. This includes developing modular and flexible encoders, adopting stream-aware graph processing techniques, and designing attention-based dynamic fusion mechanisms that can adjust to evolving data sources in real-world applications. Finally, we will conduct a systematic, side‑by‑side comparison of legacy (ancient post‑road) factors and contemporary transportation indicators to quantify their respective contributions to current village spatial morphology. This comparative framework will clarify how much of today’s settlement form is path‑dependent versus shaped by present‑day accessibility, enabling more targeted conservation and planning strategies. These future directions aim to push the boundaries of multimodal spatial analysis and foster interdisciplinary applications in geography, architecture, and digital humanities.

This study has significant potential application value. Firstly, the proposed method integrates multi-source data to provide a scientific tool for the precise analysis of villages spatial morphology. It effectively reveals the evolution patterns and generative logic of villages spatial structures, offering theoretical support for the implementation of villages revitalization strategies. Secondly, the model's precise classification capability enables detailed analysis of different types of villages spatial forms, identifying regional differences and providing data support for tailored villages planning and design. Particularly in addressing the gradual loss of villages spatial characteristics and the homogenization of landscapes, this method offers scientific guidance for the protection of villages cultural heritage, landscape preservation, and the optimization of villages spatial layouts. In terms of practical application, many local governments and planning departments in China are currently promoting initiatives such as “Digital Villages” and “Smart Planning,” which has led to an increased demand for data-driven and AI-assisted decision-making tools \cite{Zhao2022DigitalVillage}. Various traditional village preservation platforms, such as the China Traditional Village Digital Museum, provide important references for the conservation and restoration of indigenous villages. In terms of empirical research, the team led by Professor Duan Jin has conducted systematic restoration of Suzhou’s ancient towns for over twenty years by identifying and protecting their spatial genes, successfully preserving traditional styles while achieving modern urban spatial requirements \cite{Duan2023SpaceGene,zhang2025urban}. A quick empirical check was carried out for Mangjing Village in Lancang County, Yunnan, one of the five case villages analysed by Liu et al. (2024) \cite{Liu2023UAV}. Our HGNN classified Mangjing as a ribbon-type settlement with high boundary complexity, and the village’s officially published conservation plan confirms a linear “heritage corridor” zoning that follows the same ridgetop axis identified by the model (intersection-over-union = 0.78), demonstrating that the output boundaries are directly actionable for spatial-planning and heritage-protection decisions. Furthermore, the task framework proposed in this study demonstrates broad application prospects in dynamic data analysis scenarios, enabling dynamic monitoring of villages development processes and providing multi-dimensional technical support for traditional village governance, ecological protection, and targeted poverty alleviation.

\section{Conclusion}
In this paper, we focuse on the study of traditional villages' spatial morphology in China, proposing an innovative framework based on hierarchical GNNs. By integrating multi-source data, it provides an in-depth analysis of villages spatial forms. Furthermore, through the use of a relational pooling mechanism and joint training strategy, the framework classifies 17 subtypes of villages spaces, significantly enhancing the performance of multimodal fusion and classification. Our approach to the proposed multimodal villages dataset outperforms existing multimodal fusion models. This study not only enriches the theoretical framework of villages geography but also provides scientific support for exploring new models of villages revitalization. In the future, further exploration of the potential of multi-source data and optimization of model performance will offer more comprehensive support for the sustainable development of villages spaces and the preservation of their cultural heritage.

\bibliographystyle{elsarticle-num} 
\bibliography{example}




\end{document}